\documentclass[11pt]{article}

\usepackage[final]{acl}

\usepackage{times}
\usepackage{latexsym}

\usepackage[T1]{fontenc}

\usepackage[utf8]{inputenc}

\usepackage{microtype}

\usepackage{inconsolata}

\usepackage{graphicx}

\usepackage{amsmath, mathtools, amssymb}
\usepackage{tabularx}
\usepackage{subcaption}
\usepackage{caption}
\usepackage{fancyvrb}
\usepackage{listings}
\usepackage{url}
\usepackage{enumitem}
\usepackage{array, booktabs, multirow, makecell}
\newcolumntype{C}[1]{>{\centering\arraybackslash}p{#1}}
\newcolumntype{L}[1]{>{\arraybackslash}p{#1}}
\usepackage{xcolor,colortbl}
\usepackage{listings}
\usepackage{hyperref}

\usepackage{pifont}
\usepackage{framed}
\usepackage{capt-of}

\definecolor{OliveGreen}{rgb}{0,0.6,0}
\definecolor{CornellRed}{rgb}{0.7, 0.11, 0.11}
\lstdefinestyle{customJava}{
    language=Java,
    moredelim=[is][\color{OliveGreen}]{\$}{\$}, 
    moredelim=[is][\color{CornellRed}]{@}{@}      
}

\usepackage{xcolor}
\usepackage{multirow}
\usepackage{longtable}

\newcommand{\rgcolor}[1]{%
  \ifnum #1<50
    red!\number\numexpr 100-2*#1\relax!yellow%
  \else
    yellow!\number\numexpr 200-2*#1\relax!green%
  \fi
}

\newcommand{\masterybox}[2]{%
  \begingroup
  \setlength{\fboxsep}{2pt}%
  \colorbox{\rgcolor{#2}}{%
    \makebox[1.5em][c]{%
      \rule[-0.6em]{0pt}{1.5em}%
      \scriptsize #1%
    }%
  }%
  \endgroup
}

\usepackage{xspace}

\newcommand{\kaser}{\textsc{KASER}\xspace}

\title{\kaser: Knowledge-Aligned Student Error Simulator \\ for Open-Ended Coding Tasks}

\author{
Zhangqi Duan \\
University of Massachusetts \\Amherst \\
\texttt{zduan@umass.edu}
\And
Nigel Fernandez \\
University of Massachusetts \\Amherst \\
\texttt{nigel@umass.edu}
\And
Andrew Lan \\
University of Massachusetts \\Amherst \\
\texttt{andrewlan@umass.edu}
}

\begin{document}
\maketitle

\begin{abstract}
Open-ended tasks, such as coding problems that are common in computer science education, provide detailed insights into student knowledge. However, training large language models (LLMs) to simulate and predict possible student errors in their responses to these problems can be challenging: they often suffer from mode collapse and fail to fully capture the diversity in syntax, style, and solution approach in student responses. In this work, we present \kaser ({\textbf{K}nowledge-\textbf{A}ligned \textbf{S}tudent \textbf{E}rror Simulato\textbf{r}}), a novel approach that aligns errors with \emph{student knowledge}. We propose a training method based on reinforcement learning using a hybrid reward that reflects three aspects of student code prediction: i) code similarity to the ground-truth, ii) error matching, and iii) code prediction diversity. On two real-world datasets, we perform two levels of evaluation and show that: At the per-student-problem pair level, our method outperforms baselines on code and error prediction; at the per-problem level, our method outperforms baselines on error coverage and simulated code diversity. 
\end{abstract}


\section{Introduction}
Erroneous responses that students generate to problems, especially open-ended tasks such as coding that are common in computer science education, provide detailed insights into student knowledge \cite{hoq2024towards}. However, understanding and even simulating student errors in these tasks presents a unique challenge to large language models (LLMs): students make errors that reflect ingrained misconceptions \cite{brown,feldman} or lack of sufficient knowledge \cite{anderson}, which is different from the usual clean, expert-generated content that LLMs are pre-trained on. Another challenge is the high diversity among student responses: student codes differ significantly in syntax, style, and solution approaches. This diversity is evident even among correct responses and is more amplified among incorrect responses. Capturing them can be highly beneficial in education, enabling teachers to accurately diagnose student errors and knowledge deficiencies~\cite {diana2017instructor} and deliver personalized feedback to students~\cite{shaka-etal-2024-error}. See Section~\ref{sec:rw} for broader coverage of related work. 

A line of notable recent work has studied simulating student errors in open-ended coding tasks, focusing on analyzing sequences of student submissions as they iteratively submit code and receive feedback from the compiler~\cite{okt,miroyan2025parastudent,ross2025modeling}. Despite being able to simulate student code submissions, there are obvious limitations. First, the open-ended knowledge tracing (KT) work in~\citet{okt} uses a student model, KT~\cite{corbett1994knowledge}, to track how student knowledge evolves over time and inform an LLM to generate predictions of student code submissions. This method explicitly uses knowledge to steer LLM output but does not analyze errors. More importantly, we found that such a supervised fine-tuning (SFT)-based method suffers from mode collapse: the resulting code generation lacks diversity. When using recent LLMs, such as Qwen2.5-Coder as the backbone (their work used GPT-2), SFT on real student code 
often still results in the model generating correct code.

Second, the ParaStudent work in \citet{miroyan2025parastudent} proposed using representative code from a student's past code submissions to other problems, at various stages of attempting a problem, to help predict their future code submissions. This method successfully simulates how students progress over time and captures their errors. However, the model lacks an explicit link to student knowledge, which is key to being pedagogically useful in real-world education scenarios. Moreover, their error evaluation is conducted at a population level (simulating an overall error distribution across students), rather than at a more challenging individual student level (simulating which specific errors a student will make). 
Third, the work in~\citet{ross2025modeling} proposed to use prior code submissions as ``chain-of-thought''~\cite{wei2022chain} to simulate the student's final submission to a coding task. This method successfully captures students' coding style and efficiency, but only uses student IDs to capture these properties, without explicitly modeling student knowledge. Neither of these methods studies student errors in detail; the former broadly categorizes them into logical and runtime errors, and the latter does not use errors in its evaluation. 


\subsection{Contributions}

In this work, we present \kaser ({\textbf{K}nowledge-\textbf{A}ligned \textbf{S}tudent \textbf{E}rror Simulato\textbf{r}}), a method for open-ended student response simulation (we will publicly release our code) with one core goal: aligning student errors with their knowledge on a set of knowledge components (KCs). We ground our work on two real-world student coding datasets and summarize our contributions as:
\begin{enumerate}
    \item We develop a group-relative policy optimization (GRPO)-based method \footnote{Our code can be found at: \url{https://github.com/umass-ml4ed/code_personalization}.} to align errors in output student code with input student knowledge. We construct a three-part reward function; in addition to a standard similarity reward between predicted and ground-truth student code, we add two novel parts: First, a \textit{group-level reward} that encourages diversity in student code predictions, to prevent the model from mode collapse and capture high diversity among student-written code. Second, an error-overlap reward that reflects how errors (if any) present in predicted student code match those (if any) present in actual student code. This reward goes beyond surface code similarity metrics and captures student errors, aligning them with student knowledge levels in an interpretable way. 
    \item We conduct extensive quantitative evaluation at two levels: per-student-problem pair and per-problem. Results on the former show that our method outperforms baselines in terms of student code prediction accuracy, especially on anticipating errors. Results on the latter show that our method anticipates a wider range of student errors in an unseen problem. We also qualitatively show how error prediction changes for different knowledge levels, which can be useful to provide diagnostics and feedback to instructors and students. 
\end{enumerate}


\section{Task Setup}



In open-ended coding tasks that are common in computer science education, students are usually asked to write open-ended code to implement a function according to problem specifications; see the top left part of Figure~\ref{fig:model} for an illustrative example. We denote each problem as $p$. In student modeling literature, each problem is characterized as testing students' mastery of a set of knowledge components (KCs), which we denote as $W$. Our goal is to predict the code written by the student in response to the problem, which we denote as $c$, and in particular, the set of errors (if any) contained in the code, which we denote as $E$. 


\subsection{Error Annotation}

Given a student-written code for a problem, we use an automatic error annotation pipeline based on OpenAI's o4-mini~\cite{o4mini} reasoning model. We resort to this approach for two reasons: First, many codes written by actual students may contain issues, such as infinite loops, that cannot be properly executed. Moreover, many coding tasks do not come with test cases, as is the case with the two datasets we work with; see Section~\ref{sec:experiments} for details. Therefore, an LLM prompting approach is widely applicable. 
Second, compilers may not be able to identify logical errors beyond surface-level syntax and runtime errors, which are common in student codes~\cite{hoq2024towards}. 
Specifically, we use a three-step approach inspired by~\cite{duan2025automated}: 1) generating error annotations for each student code-problem pair independently through prompting, 2) clustering errors across all pairs, and 3) summarizing each cluster to obtain a representative error description. We detail each step below. 

\paragraph{Error Generation}
For each student code-problem $(c,p)$ pair, we prompt o4-mini in a chain-of-thought manner to generate a list of errors $E$ in the code submission covering syntax, runtime, and logical errors. We include representative examples of errors in all three categories, chosen from a list of frequent errors that novice programmers make~\cite{altadmri201537,zhou2021learning}.
We instruct the model to provide a concise reasoning behind identifying a particular error, followed by the error category and its standardized description.

\paragraph{Clustering Errors}
To aggregate similar errors across different student codes under the same problem and also across problems, we cluster error descriptions by first computing the Sentence-BERT~\cite{reimers-2019-sentence-bert} embedding of their textual descriptions, followed by applying Hierarchical Agglomerative Clustering~\cite{mullner2011modern}, using cosine similarity as the distance function. We can adjust the number of clusters to control the abstraction level of the error descriptions, yielding descriptions that are informative yet generalizable across student codes.

\paragraph{Representative Labels for Error Clusters}
In the final step, we prompt o4-mini with chain-of-thought reasoning to generate a single representative error description for each cluster. The model first identifies an error that reflects the majority of errors in the cluster, then abstracts it by removing problem-specific details to produce a generalized description. We use these representative errors to annotate all student-code pairs by mapping initial annotations to their corresponding cluster labels.
We include all prompts in Appendix~\ref{apdx:prompts}. 

We conduct two human evaluations to assess the reliability of using an LLM-as-a-judge to identify and classify errors in code. In the first evaluation, we randomly select $50$ problems from the FalconCode dataset and analyze both ground-truth and generated code for each problem. Human annotators are asked to perform the same task as the LLM: selecting errors in the code from a predefined error list. Results show an average F1 score between the LLM and human label of $0.749$, indicating substantial agreement, and inter-rater F1 score is $0.814$. To further analyze performance across error categories, we report stratified human evaluation results: the average F1 is $0.830$ for syntax errors, $0.893$ for runtime errors, and $0.917$ for logical errors.

In the second evaluation, we assess the accuracy of error type classification by measuring the Jaccard agreement between LLM-predicted and human-annotated error types for identified errors on $50$ randomly sampled generated codes. The average agreement is $0.800$ for syntax errors, $0.778$ for runtime errors, and $0.872$ for logical errors. These results further demonstrate that our judge model can not only reliably identify errors but also accurately classify their types, despite the non-trivial task of choosing among $10$ possible errors per problem (see Appendix~\ref{apdx:prompts} for details).


\subsection{Student Error Simulation}

Our ultimate goal is to anticipate errors students might make in their code when responding to a problem. Given a history of prior student codes, where we define each response as $x_t \coloneq (p_t, W_t, c_t, E_t)$, with $t$ indexing time steps that indicate the order in which each student attempts problems. Therefore, given $x_0, \ldots, x_t$, our goal is to predict their code submission, $c_{t+1}$, and especially the errors contained in it, $E_{t+1}$, submitted by the student to a future problem $p_{t+1}$. 


\section{Methodology}

We now detail \kaser's student code/error-knowledge alignment method via RL. 

\begin{figure*}
\includegraphics[width=\textwidth]{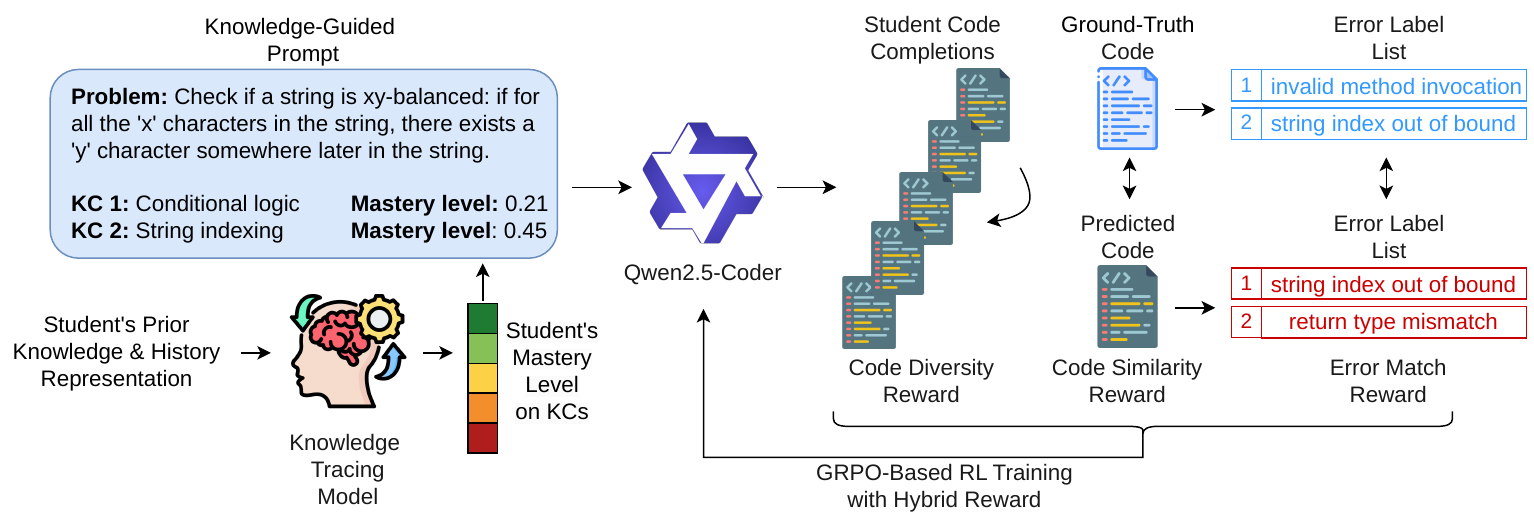}
\vspace{-.5cm}
\caption{\kaser estimates an interpretable student knowledge profile and trains an LLM via GRPO with a hybrid reward to simulate knowledge-aligned errors in predicted student code.} 
\label{fig:model}
\vspace{-.3cm}
\end{figure*}


\subsection{Student Knowledge Estimator}

\kaser first trains a knowledge estimator (KE) that transforms a student's history of submitted code into a knowledge profile of mastery over KCs. We use a knowledge tracing (KT) model to estimate a student's $d$-dimensional knowledge state vector $h_t \in \mathbb{R}^{d}$ based on the code they wrote for all prior problems. 
We compress this knowledge state $h_t$ into a k-dimensional mastery vector $m_t \in [0,1]^k$, where k denotes the total number of unique KCs, by passing it through a linear layer with weights $W_m \in \mathbb{R}^{k\times d}$ and bias $b_m \in \mathbb{R}^k$, followed by a sigmoid activation to map the values of $m_t$ to be in the range of $[0,1]$, given by $m_t = \sigma(W_m h_t + b_m)$. Each dimension of $m_t$ denotes the student's \textit{interpretable} mastery level on the $j$-th unique KC, with higher values indicating higher mastery.

We use a compensatory model~\cite{maier2021challenges} which takes the average of individual student KC mastery levels to obtain an overall mastery level
$\hat{y}_{t+1} = \frac{1}{\sum_{k=1}^K \mathbb{I}(w_k)}\sum_{k=1}^{K} m_t^k\cdot \mathbb{I}(w_k)$, where the indicator function $\mathbb{I}(w_k)$ is $1$ if the KC $w_k$ is associated with the problem, and $0$ otherwise. To train this model, 
we minimize the BCE loss, which for one student response is given by:
\begin{equation*}
    \mathcal{L}_{\text{KE}} = a_{t+1} \cdot \log \hat{y}_{t+1} + (1-a_{t+1})\cdot \log (1-\hat{y}_{t+1}),
\end{equation*}
where $a_{t+1}$ denotes the ground-truth binary-valued submission correctness.
We average this loss across all students' code submissions to all problems.


\subsection{Student Code Predictor}
\label{sec:student_code_predictor}
\kaser trains a large language model (LLM), specifically Qwen2.5-Coder 7B Instruct~\cite{hui2024qwen2}, via supervised finetuning (SFT), for student code prediction (CP) on the next coding problem, given their knowledge profile.
We construct our LLM prompt by including both the textual problem statement and the student's mastery level on the KCs associated with the problem. Our knowledge-guided prompt is given by:
\texttt{Problem: {$p_t$}. KC $1$: <$w^1$>. The student's mastery level on <$w^1$> is: $m_t^1$. KC $2$: <$w^2$>$ \ldots$}, as shown in Figure~\ref{fig:model} and Table~\ref{apdx: kaser_prompt}. Here, $m_t^1 \in [0,1]$ denotes the student's mastery level on KC $w^1$ as a real-valued number obtained from the mastery vector $m_t$ estimated by the KE model.
We then prompt Qwen2.5-Coder 7B Instruct to generate the predicted code $\hat{c}$ token-by-token. We minimize the token-level cross-entropy loss, which for one student code is given by:
\begin{equation*} 
    \mathcal{L}_{\text{CP}} = \textstyle\sum_{n=1}^N -\log P \left( \hat{c}^n\ \middle \vert\ p,j,\{\hat{c}^{n'}\}_{n'=1}^{n-1}\right),
\end{equation*}
where N is the number of tokens in the student code. We average this loss across all students' code submissions to all problems.


\subsection{Knowledge-Aligned Student Error Simulation via RL}
We use GRPO~\cite{shao2024deepseekmath}, an RL algorithm designed to train LLMs with group-level normalized rewards, to train our student code predictor LLM to generate student codes with errors (if any) aligned with the input student knowledge profile.
In each GRPO iteration, for each input problem $p$ and student knowledge mastery profile $m$, we generate a group of $G$ candidate student codes $\{\hat{c}_1, \hat{c}_2, \ldots, \hat{c}_G\}$ from the current (old) student code predictor model $\pi_{\theta_\text{old}}$.
%
We assign a scalar reward $r_i=R(\hat{c}_i)$ to each candidate code using a hybrid reward $R$, detailed below in Section~\ref{sec:reward_design}.
Rewards $r_i$ are then z-score-normalized relative to the group to obtain the corresponding advantages $\hat{A}_i$.
%
The overall GRPO objective combines a clipped surrogate loss with a KL divergence penalty using a hyperparameter $\beta$,
$\mathcal{J}_{\text{GRPO}}(\theta)=\mathcal{L}_{\text{clip}}(\theta) -\beta D_{\mathrm{KL}}\left(\pi_\theta||\pi_{\text{ref}}\right)$,
where $\mathcal{L}_{\text{clip}}(\theta)$ follows the proximal policy optimization mechanism:
\begin{align*}
    \mathcal{L}_{\text{clip}}&(\theta) = \frac{1}{G} \sum_{i=1}^{G} \min \left\{ \frac{\pi_{\theta}\left(\hat{c}_i \mid p, m\right)}{\pi_{\theta_{\text{old}}}\left(\hat{c}_i \mid p, m\right)}\hat{A}_{i},\right.\\
    &\left.\text{clip}\left(\frac{\pi_{\theta}\left(\hat{c}_i \mid p, m\right)}{\pi_{\theta_{\text{old}}}\left(\hat{c}_i \mid p, m\right)}, 1-\epsilon, 1+\epsilon\right)\hat{A}_{i} \right\},
\end{align*}
constraining policy model updates with a clipping parameter $\epsilon$.
The KL divergence penalty regularizes the policy model $\pi_\theta$ to be close to the reference policy model $\pi_{\text{ref}}$, which is not updated. We minimize this average loss over all problem–student knowledge pairs to update the policy $\pi$.
%



\subsection{Reward Design for RL Training}
\label{sec:reward_design}
We define our reward function $R$ for RL training as a combination of three components: code similarity, error match, and code diversity. These rewards train the student simulator to generate codes that are similar to the ground-truth student code, match its errors, and encourage diversity in syntax, style, and solution approaches to prevent mode collapse. Formally, the reward $R$ is given by:
\begin{align}\label{eq:rw}
    R = R_{\text{Sim}} + R_{\text{Error}} + R_{\text{Div}},
\end{align}
where $R_{\text{Sim}}$ is the code similarity reward, $R_{\text{Error}}$ is the error match reward, and $R_{\text{Div}}$ is the code diversity reward. We detail each component below.

\paragraph{Code Similarity Reward}
We compute the similarity between the generated code $\hat{c}$ and the ground-truth student code $c$ using CodeBLEU~\cite{codebleu}, a variant of the classic text similarity metric BLEU~\cite{papineni2002bleu}. CodeBLEU measures syntactic and semantic similarity between two codes and outputs a score between $0$ and $1$, which we use as the reward $R_{\text{Sim}}=\text{CodeBLEU}(c, \hat{c})$.

\paragraph{Error Match Reward}
The code similarity reward captures surface-level semantic/syntactic similarity but may not fully reflect similarity in terms of student logic and approach, which are more important in educational applications. Therefore, we also explicitly introduce an error match reward to encourage the model to predict student code that has errors matching those (if any) in the ground-truth student code. Specifically, if both the generated and ground-truth student codes are correct, we set the reward to $1$. Otherwise, if at least one of the two codes is incorrect, we compute the overlap in their errors using the Intersection over Union (IoU) metric~\cite{everingham2010pascal}, a key metric in computer vision. We prioritize IoU over error coverage to penalize the inclusion of extraneous errors, thereby preventing reward hacking in which the model might generate excessively broken code to maximize error recall.
The error set of the predicted code is obtained by prompting a judge model, Qwen2.5-Coder 7B Instruct (see Appendix~\ref{apdx:prompts} for our prompt).
Given a set of errors $E={e}$ corresponding to the ground-truth student code $c$, and a set of errors $\hat{E}={\hat{e}}$ corresponding to the predicted code $\hat{c}$, the IoU between the predicted error set and ground-truth error set is given by:
\begin{align}
    \label{eq:iou}
    R_{\text{Error}}=\text{IoU}(\hat{E},E)=\frac{|\hat{E}\cap E|}{|\hat{E}\cup E|}.
\end{align}

\paragraph{Code Diversity Reward}
To encourage diversity in the generated student codes in syntax, style, and solution approaches, and prevent mode collapse, we use a \textit{group-level} reward which assigns a diversity score to each sampled code in a group based on its similarity with other samples, defined as:
\begin{align}
    \label{eq:code_diversity_reward}
    R_{\text{Div}}(\hat{c})=1-\max_{\hat{c_i} \in G \setminus \{\hat{c}\}}(\text{CodeBLEU}(\hat{c}, \hat{c_i})).
\end{align}
All three reward functions have ranges in $[0,1]$, putting them on the same scale. Therefore, one can also add weighting parameters to Eq.~\ref{eq:rw} to balance between the three rewards. In our experiments, we found that simply setting to equal weighting works well; we leave a more detailed study on reward function weighting to future work.


\section{Experimental Evaluation}
\label{sec:experiments}
We now detail our experimental settings for evaluating \kaser's ability to predict student code and errors (if any) from student knowledge.


\subsection{Dataset Details}

We conduct experiments on two publicly available student code datasets, CodeWorkout~\cite{codeworkout} and FalconCode~\cite{de2023falconcode}. They contain actual open-ended code submissions from undergraduate students to college-level programming problems. The former contains $246$ students, $50$ problems, $50$ KCs~\cite{duan2025automated}, and $10{,}834$ student code submissions. The latter contains $447$ students, $84$ problems, $60$ KCs, and $11{,}194$ student code submissions. Following prior work~\cite{10.1145/3706468.3706500}, we analyze only the first submission per problem, as it reflects errors arising from a student’s overall programming knowledge. Analyzing sequences of submissions to capture debugging skills is outside the scope of this work.

\begin{table*}[ht]
\centering
\small
\scalebox{1.0}{
\begin{tabular}{p{0.11\linewidth}p{0.25\linewidth}p{0.12\linewidth}p{0.13\linewidth}|p{0.11\linewidth}p{0.11\linewidth}}
\toprule
\multirow{2.5}{*}{Dataset} & \multirow{2.5}{*}{Model} & \multicolumn{2}{c}{Code Similarity} & \multicolumn{2}{c}{Error Match}\\

\cmidrule{3-6}
& & CodeBLEU@1$\uparrow$ & CodeBLEU@5$\uparrow$ & IoU@1 $\uparrow$ & IoU@5 $\uparrow$\\

\midrule
\multirow{8}{*}{\shortstack{CodeWorkout \\ (Java)}}
 & PersonaPrompt (7B) & $0.407_{\pm 0.013}$  & $0.422_{\pm 0.011}$ & $0.019_{\pm 0.019}$ & $0.026_{\pm 0.015}$ \\
 & ICL (7B) & $0.463_{\pm 0.016}$  & $0.484_{\pm 0.017}$ & $0.021_{\pm 0.015}$ & $0.027_{\pm 0.016}$ \\
 & PersonaPrompt (32B) & $0.447_{\pm 0.019}$  & $0.495_{\pm 0.015}$ & $0.053_{\pm 0.016}$ & $0.062_{\pm 0.012}$ \\
 & ICL (32B) & $0.476_{\pm 0.023}$  & $0.512_{\pm 0.017}$ & $0.081_{\pm 0.014}$ & $0.093_{\pm 0.011}$ \\
 & ParaStudent & $0.500_{\pm 0.011}$  & $0.550_{\pm 0.019}$ & $0.100_{\pm 0.014}$ & $0.198_{\pm 0.018}$ \\
 & Student SFT & $0.501_{\pm 0.014}$  & $0.565_{\pm 0.018}$ & $0.115_{\pm 0.018}$ & $0.244_{\pm 0.021}$ \\
\cmidrule{2-6}
 & \kaser w/o $R_{\text{Sim}}$ & $0.483_{\pm 0.009}$  & $0.499_{\pm 0.011}$ & $0.119_{\pm 0.012}$ & $0.232_{\pm 0.016}$ \\
 & \kaser w/o $R_{\text{Error}}$ & $0.510_{\pm 0.013}$  & $0.543_{\pm 0.010}$ & $0.100_{\pm 0.015}$ & $0.201_{\pm 0.009}$ \\
 & \kaser w/o $R_{\text{Div}}$ & $0.500_{\pm 0.021}$  & $0.541_{\pm 0.014}$ & $0.104_{\pm 0.018}$ & $0.212_{\pm 0.011}$ \\
 & \kaser (ours) & $\mathbf{0.524^{*}}_{\pm 0.013}$  & $\mathbf{0.599^{*}}_{\pm 0.022}$ & $\mathbf{0.157^{*}_{\pm 0.024}}$ & $\mathbf{0.276^{*}_{\pm 0.015}}$ \\

\midrule

\multirow{8}{*}{\shortstack{FalconCode\\ (Python)}}
 & PersonaPrompt (7B) & $0.411_{\pm 0.019}$  & $0.441_{\pm 0.015}$ & $0.033_{\pm 0.016}$ & $0.040_{\pm 0.012}$ \\
 & ICL (7B) & $0.428_{\pm 0.023}$  & $0.452_{\pm 0.017}$ & $0.039_{\pm 0.014}$ & $0.051_{\pm 0.018}$ \\
 & PersonaPrompt (32B) & $0.430_{\pm 0.014}$  & $0.460_{\pm 0.028}$ & $0.057_{\pm 0.007}$ & $0.061_{\pm 0.033}$ \\
 & ICL (32B) & $0.433_{\pm 0.017}$  & $0.459_{\pm 0.018}$ & $0.083_{\pm 0.010}$ & $0.097_{\pm 0.012}$ \\
 & ParaStudent & $0.627_{\pm 0.019}$  & $0.649_{\pm 0.015}$ & $0.142_{\pm 0.011}$ & $0.251_{\pm 0.019}$ \\
 & Student SFT & $0.642_{\pm 0.013}$  & $0.670_{\pm 0.010}$ & $0.153_{\pm 0.007}$ & $0.270_{\pm 0.015}$ \\
\cmidrule{2-6}
 & \kaser w/o $R_{\text{Sim}}$ & $0.635_{\pm 0.011}$  & $0.646_{\pm 0.021}$ & $0.146_{\pm 0.009}$ & $0.239_{\pm 0.020}$ \\
 & \kaser w/o $R_{\text{Error}}$ & $0.642_{\pm 0.015}$  & $0.661_{\pm 0.025}$ & $0.118_{\pm 0.006}$ & $0.200_{\pm 0.021}$ \\
 & \kaser w/o $R_{\text{Div}}$ & $0.640_{\pm 0.010}$  & $0.658_{\pm 0.012}$ & $0.123_{\pm 0.013}$ & $0.212_{\pm 0.011}$ \\
 & \kaser (ours) & $\mathbf{0.668^{*}_{\pm 0.017}}$  & $\mathbf{0.692^{*}_{\pm 0.015}}$ & $\mathbf{0.178^{*}_{\pm 0.008}}$ & $\mathbf{0.303^{*}_{\pm 0.016}}$ \\
\bottomrule
\end{tabular}}
\vspace{-.2cm}
\caption{Performance at the per-student-problem-pair level evaluating code similarity and error match. Best performance is in \textbf{bold}. \textsuperscript{$*$} denotes statistically significant improvement over all baselines 
($p<0.05$). 
}
\label{tab:student_level_results}
\vspace{-.3cm}
\end{table*}


\subsection{Baselines}

We compare our approach, \kaser, to state-of-the-art student code-prediction methods and several strong baselines adapted to our task. 
First, we compare with \textbf{PersonaPrompt}, which uses a persona-based prompt~\cite{he2024psychometric} containing the next problem and a persona of the student to prompt an LLM. We compress a student's history of submissions to previous problems into a persona by prompting an LLM to focus on problem-independent student problem-solving characteristics.
Second, we compare with in-context learning (\textbf{ICL}),
which follows prior work to prompt an LLM with the history of a student's most recent $k$ code submissions to previous problems as in-context examples~\cite{brown2020language} (see Appendix~\ref{apdx:prompts} for prompts used in prompting baselines). We search over $k \in [1, 5]$ and select the best performing value with $k=5$.
Third, we compare with an adapted version of \textbf{ParaStudent}~\cite{miroyan2025parastudent}, which finetunes an LLM to predict student code on the next coding problem given $k$ examples of the student's code submissions to previous problems, in an in-context tuning way~\cite{chen2022meta}. Similar to \textbf{ICL}, we use $k=5$. 
Fourth, we compare with \textbf{Student SFT}, as explained in Section~\ref{sec:student_code_predictor}, which finetunes an LLM for student code prediction on the next coding problem given their knowledge profile, following prior work that finetunes simulated student LLMs~\cite{duan2025automated}.


\subsection{Experimental Setup}
We report the average performance on $5$-fold cross-validation. We split datasets across \emph{students} (\emph{problems}) for per-student-problem-pair (per-problem) evaluation, to evaluate generalization to unseen students (problems). 
For fair comparison, we use the same base LLM as \kaser, Qwen2.5-Coder 7B Instruct, for finetuned baselines, and evaluate prompting-only baselines using both Qwen2.5-Coder 7B Instruct and Qwen2.5-Coder 32B Instruct.
We use a group size of $5$ during GRPO training.
We provide full experimental details for reproducibility in Appendix~\ref{apdx:exp_details}.


\subsection{Evaluation Metrics}

\paragraph{Student-Problem Pair Level}
At the per-student-problem pair level, following~\citet{okt}, we measure the syntactic and semantic similarity between the predicted code and the ground-truth student code using the \textbf{CodeBLEU}~\cite{codebleu} metric. 
For errors, we measure overlap between the errors (if any) present in the predicted code and the errors (if any) present in the ground-truth student code using the intersection over union (\textbf{IoU}) metric~\cite{everingham2010pascal} (see Equation~\ref{eq:iou}). 
The error set of both codes was obtained by prompting o4-mini 
(prompt in Appendix~\ref{apdx:prompts}).
To account for the high diversity of student code, we generate $K\in\{1,5\}$ codes per student-problem-pair and evaluate each against the ground-truth code, selecting the code with the best metric value.
We report averaged results across test student-problem pairs.

\begin{table}[h]
\centering
\vspace{-0.1cm}
\small
\scalebox{1}{
\begin{tabular*}{\columnwidth}{lcccc}
\hline
Model & Error Type & Precision & Recall & F1\\
\hline
Student SFT & Logical & 1.00 & 0.162 & 0.279\\
Student SFT & Runtime & 1.00 & 0.123 & 0.220 \\
Student SFT & Syntax & 1.00 & 0.069 &  0.130\\
KASER & Logical & 1.00 & 0.197 & 0.329 \\
KASER & Runtime & 1.00 & 0.130 & 0.231 \\
KASER & Syntax & 1.00 & 0.098 & 0.178 \\
\hline
\end{tabular*}
}
\vspace{-0.3cm}
\caption{Performance at the per-student-problem-pair level on error match stratified by error type.}
\label{tab:stratification_result}
\end{table}

\begin{table*}[ht]
\centering
\small
\scalebox{1.0}{
\begin{tabular}{p{0.11\linewidth}p{0.25\linewidth}p{0.12\linewidth}p{0.13\linewidth}|p{0.11\linewidth}p{0.11\linewidth}}
\toprule
\multirow{2.5}{*}{Dataset} & \multirow{2.5}{*}{Model} & \multicolumn{2}{c}{Code Diversity} & \multicolumn{2}{c}{Error Coverage}\\

\cmidrule{3-6}
& & Cos Dis $\uparrow$ & $\text{CodeBLEU}^\complement_{\text{max}}$$\uparrow$ & IoU $\uparrow$ & $\chi^2$ Dist $\downarrow$\\

\midrule
\multirow{8}{*}{\shortstack{CodeWorkout \\ (Java)}}
 & PersonaPrompt (32B) &  $0.071_{\pm 0.005}$ & $0.427_{\pm 0.019}$ & $0.310_{\pm 0.016}$ & $145.63_{\pm 4.52}$ \\
 & ICL (32B) &  $0.069_{\pm 0.003}$ & $0.411_{\pm 0.014}$ & $0.340_{\pm 0.015}$ & $139.52_{\pm 4.33}$ \\
 & ParaStudent &  $0.077_{\pm 0.002}$ & $0.476_{\pm 0.020}$ & $0.650_{\pm 0.016}$ & $114.87_{\pm 3.36}$ \\
 & Student SFT &  $0.082_{\pm 0.004}$ & $0.480_{\pm 0.015}$ & $0.700_{\pm 0.026}$ & $109.85_{\pm 3.10}$ \\
\cmidrule{2-6}
 & \kaser w/o $R_{\text{Sim}}$ & $0.081_{\pm 0.003}$  & $0.466_{\pm 0.013}$ & $0.700_{\pm 0.011}$ & $110.41_{\pm 3.25}$ \\
 & \kaser w/o $R_{\text{Error}}$ & $0.081_{\pm 0.006}$  & $0.478_{\pm 0.017}$ & $0.670_{\pm 0.008}$ & $117.19_{\pm 2.98}$ \\
 & \kaser w/o $R_{\text{Div}}$ & $0.079_{\pm 0.004}$  & $0.486_{\pm 0.012}$ & $0.680_{\pm 0.012}$ & $115.37_{\pm 4.08}$ \\
 & \kaser (ours) & $\mathbf{0.088_{\pm 0.003}}$  & $\mathbf{0.520_{\pm 0.021}}$ & $\mathbf{0.750^{*}_{\pm 0.014}}$ & $\mathbf{104.97^{*}_{\pm 3.41}}$ \\

\midrule

\multirow{8}{*}{\shortstack{FalconCode \\ (Python)}}
 & PersonaPrompt (32B) &  $0.252_{\pm 0.007}$ & $0.511_{\pm 0.015}$ & $0.298_{\pm 0.020}$ & $90.62_{\pm 6.93}$ \\
 & ICL (32B) &  $0.266_{\pm 0.003}$ & $0.508_{\pm 0.018}$ & $0.300_{\pm 0.017}$ & $87.85_{\pm 8.32}$ \\
 & ParaStudent &  $0.277_{\pm 0.009}$ & $0.597_{\pm 0.021}$ & $0.781_{\pm 0.017}$ & $53.71_{\pm 6.15}$ \\
 & Student SFT &  $0.279_{\pm 0.004}$ & $0.600_{\pm 0.022}$ & $0.755_{\pm 0.023}$ & $51.89_{\pm 5.14}$ \\
\cmidrule{2-6}
 & \kaser w/o $R_{\text{Sim}}$ & $0.275_{\pm 0.006}$  & $0.603_{\pm 0.026}$ & $0.760_{\pm 0.030}$ & $52.37_{\pm 4.67}$ \\
 & \kaser w/o $R_{\text{Error}}$ & $0.263_{\pm 0.005}$  & $0.584_{\pm 0.019}$ & $0.752_{\pm 0.021}$ & $58.55_{\pm 3.02}$ \\
 & \kaser w/o $R_{\text{Div}}$ & $0.260_{\pm 0.004}$  & $0.575_{\pm 0.016}$ & $0.757_{\pm 0.026}$ & $56.72_{\pm 4.18}$ \\
 & \kaser (ours) & $\mathbf{0.298_{\pm 0.004}}$  & $\mathbf{0.643^{*}_{\pm 0.024}}$ & $\mathbf{0.817^{*}_{\pm 0.029}}$ & $\mathbf{45.77^{*}_{\pm 5.78}}$ \\
\bottomrule
\end{tabular}}
\caption{Performance at the per-problem level evaluating code diversity and error coverage. 
}
\label{tab:problem_level_results}
\vspace{-.3cm}
\end{table*}

\paragraph{Problem Level}
At the per-problem level, we measure error coverage and code diversity. For error coverage, we measure the overlap between the set of unique errors aggregated across predicted codes and the set of errors across ground-truth student codes for a problem using \textbf{IoU}.
Taking frequency of errors into account, we also compare the two distributions of errors by calculating the \textbf{Chi-squared distance} between the error frequencies in predicted codes and in ground-truth student codes for a given problem. 
To measure code diversity, we extract embeddings of predicted student codes for a problem using the Qwen3-Embedding~\cite{qwen3embedding} model. Following~\citet{anschel2025group}, we 1) compute the average \textbf{cosine distance} between all pairs of code embeddings, capturing semantic diversity, and 2) compute the complement of the maximum CodeBLEU score ($\mathbf{\textbf{CodeBLEU}^\complement_{\text{max}}}$) between each predicted code and other predicted codes (Equation~\ref{eq:code_diversity_reward}).
We report averaged results across all test problems.


\section{Results, Analysis, and Discussion}
We now discuss our quantitative evaluation results, conduct an ablation study, and also qualitatively show that errors in predicted code are aligned with student knowledge profiles.


\subsection{Quantitative Evaluation}

\paragraph{Student-Problem Pair Level}
Table~\ref{tab:student_level_results} shows the average performance (and standard deviation) on both datasets at the per-student-problem pair level. 
On both datasets, we see that prompting-based methods perform poorly, suggesting that student-error simulation on coding tasks is inherently difficult. However, larger scale language models yields slight performance improvements under prompting-based methods. In addition, these results suggest that pretrained models exhibit an overly optimistic bias toward generating correct code, even when the student's submission history indicates low skill mastery. 
Our proposed approach, \kaser, outperforms all methods with statistical significance ($p<0.05$ under paired t-test),
including the strongest finetuning-based baseline (Student SFT), on all metrics. 

We further report error match result for K=1 on the CodeWorkout dataset stratified by error types. We compare our method, KASER, with the strongest baseline, Student SFT. From table~\ref{tab:stratification_result}, we see that both models achieve perfect precision in predicted errors, indicating that generated errors are consistently relevant. However, KASER substantially improves recall and thus F1, demonstrating broader coverage of errors, including rarer cases, consistent with the intended effect of our reward design for GRPO training. While syntax errors remain challenging for pretrained LLMs, reflected in lower recall overall, KASER still incorporates them more effectively than the baseline. Overall, these results show that KASER consistently outperforms the strongest baseline across all three error categories and evaluation metrics. In addition, we report the aggregate results across all student submissions, in Appendix~\ref{apx: add_quantitative}.

All these results confirm that aligning code and errors with an explicit, interpretable student knowledge profile is effective; without that, SFT tends to focus on surface code similarity and learns superficial patterns, not student error patterns. 
We observe low IoU values on error matching across methods. Our qualitative analysis reveals a lack of syntax errors predicted by the methods, possibly because pre-trained LLMs have a low propensity to generate syntactically invalid code. 

\begin{table*}[ht]
\small
\centering
\begingroup
\setlength{\tabcolsep}{3pt}
\begin{tabular}{p{0.09\linewidth}|c p{0.22\linewidth}|c p{0.22\linewidth}|c p{0.01\linewidth}p{0.2\linewidth}}

\toprule
\multicolumn{7}{L{15.4cm}}{Write a function in Java: 
Given 3 int values, a, b, and c, return their sum. However, if one of the values is 13 then it does not count towards the sum and values to its right do not count. So for example, if b is 13, then both b and c do not count.}\\

\midrule
\multicolumn{1}{c}{KC} & \multicolumn{1}{c}{KS} & \multicolumn{1}{c}{\kaser Pred Stud Code 1} & \multicolumn{1}{c}{KS} & \multicolumn{1}{c}{\kaser Pred Stud Code 2} & \multicolumn{1}{c}{KS} & \multicolumn{1}{c}{\kaser Pred Stud Code 3}\\ 
\midrule

\multirow{6}{*}{\shortstack{Numerical\\comparison\vspace{0.2cm} \\ Return\\statement\vspace{0.2cm} \\ Conditional\\logic\vspace{0.2cm} \\ Arithmetic\\operations}}
&
\multirow{6}{*}{
\shortstack{
\masterybox{0.12}{12}\vspace{0.12cm}\\
\masterybox{0.39}{39}\vspace{0.11cm}\\
\masterybox{0.35}{35}\vspace{0.19cm}\\
\masterybox{0.2}{20}
}} 
& 
\begin{lstlisting}
public int luckySum
    (int a, int b, int c){
    if (a == 13) {
        return c;
    }
    else if (b == 13) {
        return a;
    }
    else if (c = 13) {
        return b;
    }
    else{
        return a+b+c;
    }
}\end{lstlisting}
&
\multirow{6}{*}{
\shortstack{
\masterybox{0.68}{68}\vspace{0.12cm}\\
\masterybox{0.1}{10}\vspace{0.11cm}\\
\masterybox{0.21}{21}\vspace{0.19cm}\\
\masterybox{0.46}{46}
}} 
& 
\begin{lstlisting}
public int luckySum
    (int a, int b, int c){
    int answer = 0;
    if (a == 13) {
        answer = 0;
    }
    else if (b == 13) {
        answer = a;
    }
    else if (c == 13) {
        answer = a + b;
    }
}\end{lstlisting}
&
\multirow{6}{*}{
\shortstack{
\masterybox{0.84}{84}\vspace{0.12cm}\\
\masterybox{0.88}{88}\vspace{0.11cm}\\
\masterybox{0.76}{76}\vspace{0.19cm}\\
\masterybox{0.75}{75}
}} 
& 
\begin{lstlisting}
public int luckySum
    (int a, int b, int c){
    if (a == 13) {
        return 0;
    }
    if (b == 13) {
        return a;
    }
    if (c == 13) {
        return a + b;
    }
    else {
        return a + b + c;
    }
}\end{lstlisting}

\\

\bottomrule
\end{tabular}
\endgroup
\caption{\kaser simulates knowledge-aligned student errors (if any) in predicted codes. 
}
\label{tab:knowledge_aligned_student_errors_qual}
\end{table*}

\paragraph{Problem Level}
Table~\ref{tab:problem_level_results} shows the average performance (and standard deviation) on both datasets at the per-problem level. We show the results using larger LLM for prompting-based methods due to it is a stronger baseline.
\kaser outperforms all methods 
similar to the per-student-problem pair level. 
This result shows that when generalizing to previously unseen problems, \kaser can predict the high diversity in syntax, style, and erroneous solution approaches in student code. 
This result suggests that encouraging diversity makes \kaser better at preventing mode collapse than SFT alone. 
We provide a diversity visualization of predicted code embeddings in Appendix~\ref{apx: diversity_vis}.

\paragraph{Ablation Study}
We perform an ablation study on our reward design used in RL training on both datasets. Results are reported on the per-student-problem level in Table~\ref{tab:student_level_results} and on the per-problem level in Table~\ref{tab:problem_level_results}. 
We see that 
removing $R_{\text{Sim}}$ leads to a big drop in performance on code similarity metrics at the per-student-problem level (e.g. $16\%$ decrease on CodeBLEU@5 on CodeWorkout), as expected. 
Our error match reward $R_{\text{Error}}$ is also critical in training \kaser to predict student code with errors matching those (if any) present in the ground-truth student code; removing it leads to a big drop in performance on error matching IoU metric at the per-student-problem level (e.g. $36\%$ decrease on IoU@1 on CodeWorkout), and on the error coverage metrics at the per-problem level (e.g. $11\%$ increase on $\chi^2$ distance on CodeWorkout).
Notably, our code diversity reward $R_{\text{Div}}$ is also important in preventing mode collapse; removing it not only degrades performance at the per-student-problem level on both metrics but significantly drops performance at the per-problem level (e.g. $12\%$ decrease on cosine distance on FalconCode).



\subsection{Qualitative Evaluation}

\paragraph{\kaser simulates knowledge-aligned student errors}
We now use a qualitative case study to demonstrate how \kaser predicts different types of errors in student code aligned with their knowledge profile.
Table~\ref{tab:knowledge_aligned_student_errors_qual} shows sample predicted codes from three students with different mastery levels on KCs required to solve a problem, which involves returning the sum of three integers based on conditional logic. 
For the first student with low mastery on numerical comparison and arithmetic operations, \kaser predicts code containing an incorrect assignment operator instead of a numerical comparison operator, i.e., ``\texttt{c=13}'' instead of ``\texttt{c==13}'', as well as an incorrect arithmetic operation ``\texttt{return b}'' instead of ``\texttt{return a+b}''.
For the second student with low mastery on using return statements and conditional logic, \kaser predicts code missing a return statement as well as omitting the final else conditional block of returning the sum of all three numbers.
For the third student with high mastery on all KCs, \kaser predicts code that correctly solves the problem.
In contrast, the Student SFT method predicted code that correctly solved the problem for all three students, despite their varying knowledge levels.

\paragraph{Error Analysis}
We note that the IoU metric values are generally low for all methods in Table~\ref{tab:student_level_results}, 
around $0.1$ for a maximum possible value of $1$. To illustrate what caused it, we show an example in Appendix~\ref{apdx:error_alignment_qual}. We find that \kaser accurately predicts logical and runtime errors (``missing logic branch'') but struggles with syntax errors (``unbalanced brace''). This observation suggests that, despite extensive training on student-written code that often contains basic syntax errors, we still struggle to make pre-trained LLMs forget about code syntax and learn to make errors like students do, which highlights a key direction for future work. 



\section{Related Work}
\label{sec:rw}

There is a line of existing work~\cite{peaches,peach} on analyzing student-generated code for tasks such as error analysis and automated feedback generation that are meaningful in CS education settings. 
Program synthesis techniques have been applied for CS education to generate student code~\cite{adish,koutcheme2026teachinglanguagemodelscode}, new problems~\cite{task_synth}, and provide real-time hints~\cite{hints}.
We focus on improving student-error simulation in programming problems by aligning errors with interpretable student knowledge profiles via RL.

An increasing body of research has studied simulating students using LLMs on open-ended coding~\cite{okt}, math~\cite{feng2025reasoning,ozyurt2024automated}, and dialogue~\cite {scarlatos2025exploring} tasks, often using variants of KT~\cite{corbett1994knowledge}. While RL has achieved strong results across a wide range of domains~\cite{talpaert2019exploringapplicationsdeepreinforcement,luo2025agentperformingautonomousstock}, there is limited work on leveraging RL to train student models in educational settings; \citet{scarlatos2025smart} aligns student LLMs with instructed ability using direct preference optimization~\cite{rafailov2023direct}.
In contrast, \kaser predicts student code and errors (if any) aligned with an explicit, interpretable student knowledge profile, trained using RL with a hybrid reward.



\section{Conclusions and Future Work}

In this paper, we presented \kaser, a novel approach to improve student error simulation in predicted codes to open-ended programming problems by aligning errors with student knowledge. We proposed an RL-based training method using a hybrid reward reflecting three aspects of student code prediction: 1) code similarity to the ground-truth, 2) error matching, and 3) code prediction diversity to prevent mode collapse of the LLM.
Through extensive experiments on two real-world student code datasets covering both Java and Python, we show that \kaser outperforms other methods at both the per-student-problem-pair level on code and error prediction, as well as at the per-problem level on error coverage and predicted code diversity. We discuss potential use cases of our work in real-world CS education settings in Appendix~\ref{apdx:possible_use}.

There are many avenues for future work. 
First, we can attempt to simulate debugging errors by analyzing the changes across multiple student code submissions to the same problem.
Second, we can explore post-training techniques to incorporate syntax errors into LLM prediction.
Third, we can explore the applicability of \kaser in other domains such as math and dialogues.

\section*{Acknowledgments}
This work is partially supported by the NSF under grants 2153481, 2237676, and 2418657. 


\section*{Limitations}
We identify several technical and practical limitations of our work. First, our error annotation pipeline relies on an LLM without access to test case information. Incorporating test case results could enable more accurate error annotations and provide finer-grained error categorization, particularly for edge cases. Second, our error evaluation using the IoU metric relies on an automated LLM-as-a-judge approach, as large-scale human annotation would be costly and time-consuming. While our human evaluation shows moderate agreement between LLM-based and human-annotated error labels, broader human evaluation could yield more reliable and precise assessments.


\section*{Ethical Considerations}
There are several potential societal benefits to our work. Primarily, accurate student modeling with error simulation can greatly benefit educational assessment: it allows instructors to anticipate student difficulties on open-ended programming tasks before deployment, it enable fine-grained analysis of misconceptions in code, and it can support personalized student assistance and instructional analysis. There are several potential risks to our work as well. There is a concern that such systems could replace human educator jobs, which is a shared concern across most domains with AI applications. We emphasize that our approach is intended to support, rather than replace, educators. As bias is common in AI systems, simulated students may not sufficiently represent minority groups, thus leading to calibration errors for these populations. Future work should study these effects before deploying any simulation based system.


\bibliography{custom}

\appendix


\section{Additional Experimental Details}
\label{apdx:exp_details}
The CodeWorkout~\cite{codeworkout} dataset was introduced in the Second CSEDM Data Challenge~\cite{csedm}. It contains actual open-ended code submissions from undergraduate students in an introductory \textit{Java} programming course at a US university. The dataset contains textual problem statements and corresponding KC tags (estimated programming concepts) from a fixed set of $50$ KCs released in ~\citet{duan2025automated}. In total, $246$ students attempt $50$ problems covering various concepts. In our work, we analyze students' first submissions to each problem, leading to a total of $10{,}834$ submissions. The average code length is 40.8 tokens and 15.8 lines, with an average of 2.35 errors per incorrect submission. The FalconCode~\cite{de2023falconcode} dataset consists of actual open-ended code submissions from undergraduate students in an introductory \textit{Python} programming course at a US university. The dataset contains textual problem statements and KC tags from a predefined set of $60$ KCs released in~\citet{duan2025automated}. In total, $447$ students attempt $84$ problems in skill-based format. Similar to CodeWorkout, we analyze students' first submission to each problem, resulting in a total of $11{,}194$ code submissions with an average length of 20.3 tokens and 5.03 lines, and an average of 1.83 errors per incorrect submission.
 
We perform $5$-fold cross-validation and report the average across different test sets. We split both datasets across \emph{students} for per-student-problem pair-level evaluation and across \emph{problems} for per-problem-level evaluation, to evaluate generalization to unseen students and problems, respectively. We use a $80\%$-$10\%$-$10\%$ train-val-test split. 
For a fair comparison, we use the same base LLM as \kaser, Qwen2.5-Coder 7B Instruct, for baselines that require finetuning, and a larger model, Qwen2.5-Coder 32B Instruct, for prompting-only baselines and we use vLLM \cite{kwon2023efficientmemorymanagementlarge} for code generation. We load base LLMs with 8-bit quantization and finetune via Low-Rank Adaptation (LoRA)~\cite{hu2022lora}. We set LoRA's rank=$16$, alpha=$32$, and dropout=$0.05$. We perform a preliminary hyperparameter search for best performance for all models that require further training. For ParaStudent and Student SFT, we train for 5 epochs, with learning rate=1e-5, linear warmup for $10\%$ of steps, and batch size=$32$ using gradient accumulation. For \kaser, we continue training after SFT and set learning rate=1e-6, $\beta=0.1$, number of completions=5 for GRPO training. We use AdamW \cite{loshchilov2018decoupled} optimizer for ParaStudent, Student SFT and \kaser. Training ParaStudent and Student SFT models takes approximately $120$ minutes on FalconCode and $70$ minutes on CodeWorkout for one epoch and \kaser training converges in $1$ ($1$) epochs on CodeWorkout (FalconCode) with each epoch taking $1620$ ($960$) minutes. All experiments are conducted on a single NVIDIA L40S 48GB GPU. At inference time, we set temperature=0.7, p=1, and top\_k=40 for code generation for all models. 

For codeBLEU, we adopt the official implementation provided in the microsoft/CodeXGLUE. In addition, we use scipy library to perform the hierarchical agglomerative clustering and the embedding distance calculation. We use Qwen3-Embedding 8B to get the code embedding to calculate the cosine distance metric at problem level. To the best of our knowledge, all software and models we build our implementation on have open-source licenses or no available license. Additionally, we are within their intended terms of use of all software and with OpenAI API. If we release code, we will ensure the license and terms reflect the sources we build on.


\section{Possible Use Cases in CS Education}
\label{apdx:possible_use}
We now discuss how knowledge-aligned student code prediction with accurate error modeling can support real-world CS educational practices. By generating code that not only aligns with a student’s estimated knowledge state but also reproduces the same types of errors observed in ground-truth student submissions, our approach enables a fine-grained understanding of how specific misconceptions manifest in code. For example, if a student’s knowledge profile indicates partial mastery of conditionals and the predicted code consistently exhibits missing boundary checks or non-boolean conditions, this suggests a systematic conceptual gap rather than an incidental syntactic error. Instructors can use such error-aligned predictions to interpret student understanding through the lens of how errors occur, rather than only whether a solution is correct.

Accurate error prediction also allows instructors to anticipate student difficulties on open-ended programming tasks before deployment. Given a new assignment and a cohort’s knowledge representations, the model can simulate likely student solutions and their associated errors. This enables educators to proactively design instructional materials, such as targeted hints, scaffolding questions, or example walkthroughs, that directly address the most probable misconceptions. Compared to coarse correctness-based signals, error-aligned code prediction provides more actionable insight for adjusting problem difficulty, sequencing learning objectives, or introducing prerequisite review content.

Finally, knowledge aligned error prediction can support personalized student assistance and instructional analysis. Predicted code that reflects ground-truth error patterns can be used to evaluate and improve automatic feedback systems and existing error labeling schemes. For student support, when a learner struggles with a problem, instructors can identify other students with similar knowledge states and error patterns who later succeeded, and examine how their errors were resolved over time. These code trajectories can then be used to provide more precise and pedagogically appropriate guidance, supporting students while keeping human educators in the loop.


\begin{figure}
  \centering
  \includegraphics[width=\linewidth]{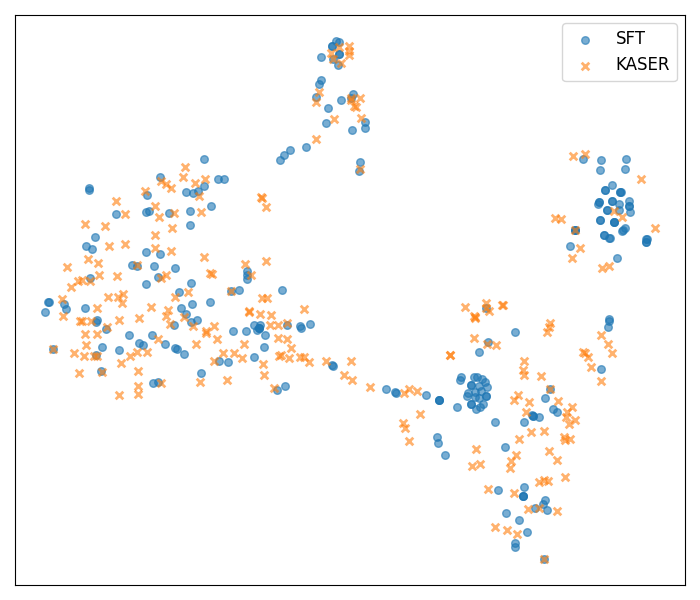}
  \caption{Diversity visualization of student codes. \kaser predicts student codes with higher diversity and alignment with ground-truth student codes compared to StudentSFT.}
  \label{fig:code_embeddings}
  \vspace{-.5cm}
\end{figure}

\section{Additional Quantitative Evaluation}
\label{apx: add_quantitative}
We report the performance at the per-student-problem pair level across all student submissions. We again compare our approach, KASER, to the strongest baseline, Student SFT. We see similar trends to the results from the first submission, with KASER outperforming the best baseline on both code similarity and error match across all metrics from Table~\ref{tab:all_submission_result}.

\setlength{\tabcolsep}{2pt}
\begin{table}[h]
\centering
\vspace{-0.1cm}
\small
\scalebox{1}{
\begin{tabular*}{\columnwidth}{lcccc}
\hline
Model & CodeBLEU@1 & CodeBLEU@5 & IoU@1 & IoU@5\\
\hline
SFT & 0.375 & 0.438 & 0.097 & 0.143\\
KASER & 0.403 & 0.466 & 0.115 & 0.179 \\
\hline
\end{tabular*}
}
\vspace{-0.3cm}
\caption{Performance at the per-student-problem-pair level across all student submissions.}
\label{tab:all_submission_result}
\end{table}

\section{Additional Qualitative Evaluation}

\subsection{Predicted Code Diversity Visualization}
\label{apx: diversity_vis}

\begin{table*}[t]
\centering
\renewcommand{\arraystretch}{0.96}
\vspace{-.1cm}
\scalebox{.97}{
\begin{tabular}{p{0.5\linewidth}|p{0.5\linewidth}}
\toprule
\multicolumn{2}{p{1\textwidth}}{Problem: Write a function in Java that implements the following logic: Your cell phone rings. Return true if you should answer it. Normally you answer, except in the morning you only answer if it is your mom calling. In all cases, if you are asleep, you do not answer.}\\
\midrule
\multicolumn{1}{c}{Ground-Truth Student Code Submission} & \multicolumn{1}{c}{Predicted Student Code Submission}\\
\midrule
\multirow{-1.8}{*}{\begin{lstlisting}[style=customJava]
public boolean answerCell(boolean isMorning, 
boolean isMom, boolean isAsleep){
    boolean answer;
    if(isMorning && isMom) {
        return true;
    }
    else if(isMorning || @isAsleep@) {
        return false;
    }
    else {
        return true;
    }
    @return answer@;
}\end{lstlisting}

} & {\begin{lstlisting}[style=customJava]
public boolean answerCell(boolean isMorning, 
boolean isMom, boolean isAsleep){
    boolean answer;
    if(isMorning && isMom) {
        answer = true;
        return answer;
    }
    else if(isMorning || @isAsleep@) {
        answer = false;
        return answer;
    }
    else {
        answer = true;
        return answer;
    }
}\end{lstlisting}

} \\
\bottomrule
\end{tabular}}
\vspace{-0.3cm}
\caption{Example showing \kaser is able to simulate the logical error in the ground-truth code, while less likely to simulate the compilation error.}
\label{tab:qualitative_error_analysis}
\end{table*}

We embed predicted student codes from \kaser and the best-performing baseline (Student SFT) into $4, 096$-dimensional code embeddings using the Qwen3-Embedding 8B~\cite{qwen3embedding} model. We visualize the embedding space using t-SNE in Figure~\ref{fig:code_embeddings}. We see that student codes predicted by \kaser occupy a greater region of the embedding space compared to SFT, showing that our method preserves diversity in student code better than baselines. This advantage can also be seen in side-by-side comparisons between predicted code examples in Appendix~\ref{apdx:code_examples_diversity}.

\subsection{Error Analysis Examples}
\label{apdx:error_alignment_qual}
We conduct an analysis on specific error types that \kaser successfully simulates and those it struggles to generate on the CodeWorkout dataset. We identify the most successful simulations by examining the intersection of generated and ground-truth errors, finding that the top three most frequently matched errors are ``incorrect comparison logic'', ``missing logic branch'', and ``array index out of bound''. In contrast, we analyze set difference to uncover the errors \kaser most frequently fails to generate: ``unbalanced brace'', ``unreachable code'', and ``type mismatch''. This distinction suggests that while our model effectively simulates logical errors, it encounters difficulty generating syntax errors, likely because the backbone models are pre-trained on vast corpora to prioritize syntactically valid code. Table~\ref{tab:qualitative_error_analysis} provides a qualitative example illustrating this distinction. In both the ground-truth and generated code, the model simulates a common student coding habit by introducing a boolean variable, and both implementations contain a logical error that checks other input values before "isAsleep", leading to an unexpected output when all three inputs are true. However, the ground-truth code includes an additional return answer statement even though all cases are already covered by the conditional logic, resulting in an unreachable code error. In contrast, \kaser does not generate this compilation error.

\subsection{Qualitative Predicted Code Examples for Diversity Visualization}
\label{apdx:code_examples_diversity}
In this section, we present side-by-side comparisons of code generated by \kaser, Student SFT, and ICL on a single problem, shown in Table~\ref{tab:qualitative_more_example} for CodeWorkout and Table~\ref{tab:qualitative_more_example_falcon} for FalconCode. As shown in Table~\ref{tab:qualitative_more_example}, the ground-truth student code contains a logical error: the code immediately returns 10 even if all three inputs are the same. In the outputs generated by \kaser, different completions exhibit varied implementations, yet all consistently reproduce the same logical error as the ground-truth code. This indicates that \kaser is able to generate more diverse code while remaining aligned with the student’s knowledge level. In contrast, both Student SFT and ICL produce identical code that correctly solves the problem, failing to simulate the student error.

We further present an example from FalconCode in Table~\ref{tab:qualitative_more_example_falcon}. In the ground-truth code, the student incorrectly treats body\_aches and loss\_of\_smell as boolean variables, although they are strings; consequently, even the value 'no' is evaluated as True in Python. Additionally, the student forgets to print the final result. In the \kaser generations, two completions successfully capture this logical error. Student SFT also reproduces the same error, but generates identical code across completions. In contrast, ICL again fails to simulate the student error and repeatedly produces the same correct solution. Together, these examples demonstrate that \kaser is effective at simulating more diverse code while aligning with student knowledge level.

\begin{table*}[p]
\centering
\renewcommand{\arraystretch}{1.2}
\begin{tabular}{p{0.7\textwidth} p{0.27\textwidth}}
\toprule
\textbf{CodeWorkout Problem} \\
\midrule
\multicolumn{2}{p{\textwidth}}{
You have a green lottery ticket, with ints $a$, $b$, and $c$ on it.
If the numbers are all different from each other, return $0$.
If all of the numbers are the same, return $20$.
If two of the numbers are the same, return $10$.
}
\\
\midrule
\textbf{Code} & \textbf{Error Labels} \\
\midrule
\multicolumn{2}{c}{\textbf{Human Student}} \\
\midrule

\begin{lstlisting}
public int greenTicket(int a, int b, int c)
{
    if(a == b || b==c){
        return 10;
    }else if(a ==b && b ==c){
        return 20;
    }else{
        return 0;
    }
}
\end{lstlisting}
&
[incorrect conditional logic]
\\\midrule
\multicolumn{2}{c}{\textbf{\kaser}} \\
\midrule
\begin{lstlisting}
public int greenTicket(int a, int b, int c)
{
    if(a == b || b==c){
        return 10;
    }else if(a ==b && b ==c){
        return 20;
    }else{
        return 0;
    }
}
\end{lstlisting}
&
[incorrect conditional logic] \\
\begin{lstlisting}
public int greenTicket(int a, int b, int c)
{
    if(a == b && b==c) {
        return 20;
    } else if (a!=b && b!=c) {
        return 0;
    } else {
        return 10;
    }
}
\end{lstlisting}
&
[incorrect conditional logic] \\
\midrule
\multicolumn{2}{c}{\textbf{Student SFT}} \\
\midrule
\begin{lstlisting}
public int greenTicket(int a, int b, int c)
{
    if(a == b && b == c)
                return 20;
        if((a == b || a==c) || b==c)
                return 10;
        else
        return 0;
}

\end{lstlisting}
&
[] \\
\begin{lstlisting}
public int greenTicket(int a, int b, int c)
{
    if(a == b && b == c)
                return 20;
        if((a == b || a==c) || b==c)
                return 10;
        else
        return 0;
}

\end{lstlisting}
&
[] \\
\midrule
\multicolumn{2}{c}{\textbf{ICL}} \\
\midrule
\begin{lstlisting}
public int greenTicket(int a, int b, int c) {
    if (a == b && b == c) {
        return 20;
    } else if (a == b || b == c || a == c) {
        return 10;
    } else {
        return 0;
    }
}
\end{lstlisting}
&
[] \\
\begin{lstlisting}
public int greenTicket(int a, int b, int c) {
    if (a == b && b == c) {
        return 20;
    } else if (a == b || b == c || a == c) {
        return 10;
    } else {
        return 0;
    }
}
\end{lstlisting}
&
[] \\
\bottomrule
\end{tabular}
\caption{Example code from different models for a problem from CodeWorkout dataset}
\label{tab:qualitative_more_example}
\end{table*}

\begin{table*}[p]
\centering
\renewcommand{\arraystretch}{1.2}
\begin{tabular}{p{0.7\textwidth} p{0.27\textwidth}}
\toprule
\textbf{FalconCode Problem} \\
\midrule
\multicolumn{2}{p{\textwidth}}{
Get the following values from the user: Body temperature (in Fahrenheit), Body aches (a string, which will be either "yes" or "no"), Loss of smell (a string, which will be either "yes" or "no")

Using the following rules as a guide, have your program determine a person's risk of having an infectious disease:
If the person's temperature is above 100.4 degrees, print "High Risk".
If the person is experiencing both body aches and loss of smell, print "High Risk".
For all other situations, print "Low Risk".
}
\\
\midrule
\textbf{Code} & \textbf{Error Labels} \\
\midrule
\multicolumn{2}{c}{\textbf{Human Student}} \\
\midrule

\begin{lstlisting}
temperature = float(input("What is your temperature in degrees F?\n"))
body_aches = input("Have you had body aches?\n")
loss_of_smell = input("Have you experienced a loss of smell?\n")
if temperature> 100.4:
    status=("High Risk")
elif body_aches and loss_of_smell:
    status=("High Risk")
else:
    status=("Low Risk")
\end{lstlisting}
&
[incorrect conditional logic, output missing]
\\\midrule
\multicolumn{2}{c}{\textbf{\kaser}} \\
\midrule
\begin{lstlisting}
temperature = float(input("What is your temperature in degrees F?\n"))
body_aches = input("Have you had body aches?\n")
loss_of_smell = input("Have you experienced a loss of smell?\n")
if temperature > 100.4 :
    print ("High Risk")
elif body_aches == yes & loss_of_smell:
    print ("High Risk")
else :
    print ("Low Risk")

\end{lstlisting}
&
[incorrect conditional logic] \\
\begin{lstlisting}
temperature = float(input("What is your temperature in degrees F?\n"))
body_aches = input("Have you had body aches?\n")
loss_of_smell = input("Have you experienced a loss of smell?\n")
if temperature > 100.4 and body_aches == 'yes' and loss_of_smell == 'yes':
    print('High Risk')
else:
    print('Low Risk')
\end{lstlisting}
&
[incorrect conditional logic] \\
\midrule
\multicolumn{2}{c}{\textbf{Student SFT}} \\
\midrule
\begin{lstlisting}
temperature = float(input("What is your temperature in degrees F?\n"))
body_aches = input("Have you had body aches?\n")
loss_of_smell = input("Have you experienced a loss of smell?\n")
if temperature > 100.4 :
    print ("High Risk")
elif body_aches == yes and loss_of_smell == yes :
    print ("High Risk")
else :
    print ("Low Risk")
\end{lstlisting}
&
[incorrect conditional logic] \\
\begin{lstlisting}
temperature = float(input("What is your temperature in degrees F?\n"))
body_aches = input("Have you had body aches?\n")
loss_of_smell = input("Have you experienced a loss of smell?\n")
if temperature > 100.4 :
    print ("High Risk")
elif body_aches == yes and loss_of_smell == yes :
    print ("High Risk")
else :
    print ("Low Risk")
\end{lstlisting}
&
[incorrect conditional logic] \\
\midrule
\multicolumn{2}{c}{\textbf{ICL}} \\
\midrule
\begin{lstlisting}
temperature = float(input("What is your temperature in degrees F?\n"))
body_aches = input("Have you had body aches?\n")
loss_of_smell = input("Have you experienced a loss of smell?\n")
if temperature>100.4:
    print('High Risk')
elif body_aches == 'yes' and loss_of_smell == 'yes':
    print('High Risk')
else:
    print('Low Risk')
\end{lstlisting}
&
[] \\
\begin{lstlisting}
temperature = float(input("What is your temperature in degrees F?\n"))
body_aches = input("Have you had body aches?\n")
loss_of_smell = input("Have you experienced a loss of smell?\n")
if temperature>100.4:
    print('High Risk')
elif body_aches == 'yes' and loss_of_smell == 'yes':
    print('High Risk')
else:
    print('Low Risk')
\end{lstlisting}
&
[] \\
\bottomrule
\end{tabular}
\caption{Example code from different models for a problem from FalconCode dataset}
\label{tab:qualitative_more_example_falcon}
\end{table*}


\section{Prompts}
\label{apdx:prompts}

\subsection{Prompt for Ground-Truth Error Labeling Pipeline}
We show an example prompt used for the error labeling in Table \ref{tab:error_generation_prompt} and the prompt used for cluster summarization in Table \ref{tab:error_summarize_prompt}.

\subsection{Prompt for Models}
We show the prompt used for ICL baseline in \ref{tab: ICL_baseline_prompt}, the prompt used for persona generation in Table \ref{tab: persona_generation_prompt}, and prompt used for Persona Prompting baseline in Table \ref{tab: persona_baseline}. We also show the prompt used for Student SFT and \kaser in Table \ref{apdx: kaser_prompt}

\subsection{Prompt for LLM-as-a-Judge and Human Evaluation}
We show the prompt used for judge model to select the errors in generated code during GRPO training, which is also the prompt for IoU evaluation in Table \ref{tab:human_eval}.

\begin{table*}[h]
\centering
\begin{tabular}{p{0.95\linewidth}}
\toprule
\textbf{System Message:} \\
You are an expert Java programming instructor and automated code reviewer. \\
Given a Java programming problem and a student's buggy code solution, your task is to identify all errors present in the code. There is at least one error in the code. Use concise and standardized label/taxonomy for each error. Make sure the error label is generalizable without problem specific description. \\
Take the following error label examples as reference: \\
Syntax Error (Examples): Confusing assignment with equality, Unbalanced parentheses, Semicolon errors \\
Runtime Error (Examples): Uninitialized Variables, Parameter confusion, NullPointerExceptions \\
Logical Error (Examples): Off-by-one errors, Integer Division, Infinite Loops \\ \\

Return a JSON object with this template: \\
$\{$
"errors": [
    {
    "Reasoning": "<one sentence explanation of the error in the code>",
    "Category": "Syntax | Runtime | Logical",
    "Label": "<error label>"
    }
    ]
    $\}$\\ \\

\textbf{User prompt:} \\
Problem: A sandwich is two pieces of bread with something in between. Write a Java method that takes in a string \texttt{str} and returns the string that is between the first and last appearance of \texttt{"bread"} in \texttt{str}. Return the empty string \texttt{""} if there are not two pieces of bread.\\ \\

Code: \\
\begin{lstlisting}
public String getSandwich(String str){
    String bread = "bread";
    if (str.contains(bread) && str.length() >= 10){
        int first = str.indexOf(bread);
        int last = str.lastIndexOf(bread);
        String between = str.substring(first + 5, last);
        return between;
    }
}
\end{lstlisting} \\
\bottomrule
\end{tabular}
\caption{Example prompt for Error Labeling with In-Context Examples for different error categories}
\label{tab:error_generation_prompt}
\end{table*}

\begin{table*}[ht]
\centering
\begin{tabular}{p{0.95\linewidth}}
\toprule
\textbf{System Message:} \\
You are an experienced computer science teacher. \\
You will be provided with a list of errors from student code that refer to the same underlying errors but may vary in wording.

Your task is to:
1. Carefully examine all the errors in the list to ensure none are overlooked.
2. Reason explicitly the error refer to the same underlying concept or if they are related but represent distinct or complementary aspects of a broader theme.
3. Based on your reasoning: select one error from the list that best represents the group — choose the one that is most clearly worded, generalizable, and inclusive of the others. Remove all problem specific description from the selected error. 

Return output strictly in the following JSON format: \\
$\{$
"Reasoning": "<Exactly one sentence explaining your reasoning on the majority error>",
"Representative\_error": "<Error name>"
$\}$\\ \\

\textbf{User prompt:} \\
The error list is: ["incorrect conditional structure", "conditional logic error", "missing conditional logic", "incorrect conditional logic"] \\ \\

Now follow the instructions in system message and perform the task. \\
\bottomrule
\end{tabular}
\caption{Example prompt for Error cluster summarization}
\label{tab:error_summarize_prompt}
\end{table*}

\begin{table*}[ht]
\centering
\begin{tabular}{p{0.95\linewidth}}
\toprule
\textbf{System Message:} \\
You are a student code simulator. \\
Given a Java programming problem and the student's past code submissions, simulate the code the student would write for the given problem. Output only the code, with no explanations or comments.\\ \\

\textbf{User prompt:} \\
Past Submissions: \\ 
$\{$Past Code Submissions$\}$ \\ \\

Problem:\\
$\{$Problem Statement$\}$ \\ \\

Simulate the student written code:\\
\bottomrule
\end{tabular}
\caption{Prompt for ICL Baseline}
\label{tab: ICL_baseline_prompt}
\end{table*}

\begin{table*}[ht]
\centering
\begin{tabular}{p{0.95\linewidth}}
\toprule
\textbf{System Message:} \\
Given a student's past code submissions, generate a persona that describes the student's overall programming skills and style but not specifically to any particular problem. Output the persona description in 3-5 sentences.\\ \\

\textbf{User prompt:} \\
Past Submissions: \\ 
$\{$Past Code Submissions$\}$ \\ \\

Generate the student's programming persona:\\
\bottomrule
\end{tabular}
\caption{Prompt for persona generation}
\label{tab: persona_generation_prompt}
\end{table*}

\begin{table*}[ht]
\centering
\begin{tabular}{p{0.95\linewidth}}
\toprule
\textbf{System Message:} \\
You are a student code simulator. \\
Given a Java programming problem and the student's persona on their programming skills, simulate the code that student would write for the given problem. Output only the code, with no explanations or comments. \\ \\

\textbf{User prompt:} \\
Student Persona: \\ 
$\{$Persona Description$\}$ \\ \\

Problem:\\
$\{$Problem Statement$\}$ \\ \\

Simulate the student written code:\\
\bottomrule
\end{tabular}
\caption{Prompt for Persona prompting baseline}
\label{tab: persona_baseline}
\end{table*}

\begin{table*}[ht]
\centering
\begin{tabular}{p{0.95\linewidth}}
\toprule
\textbf{System Message:} \\
You are a student code simulator. \\
Given a programming problem and the student's mastery levels for specific knowledge components (KCs), generate $\{$language$\}$ code that reflects the understanding, including plausible student errors. Output only the code, with no explanations or comments. \\ \\

\textbf{User prompt:} \\
Problem:\\
$\{$Problem Statement$\}$ \\ \\

Student information: \\ 
KC 1: $\{$KC 1 name$\}$. The student's mastery level on $\{$KC 1 name$\}$ is $\{$KC 1 mastery level$\}$.\\
KC 2: $\{$KC 2 name$\}$. The student's mastery level on $\{$KC 2 name$\}$ is $\{$KC 2 mastery level$\}$. \\ \\

Simulate the student written code:\\
\bottomrule
\end{tabular}
\caption{Prompt for Student SFT and \kaser training}
\label{apdx: kaser_prompt}
\end{table*}

\begin{table*}[h]
\centering
\begin{tabular}{p{0.95\linewidth}}
\toprule
\textbf{System Message:} \\
You are an experienced code reviewer. \\
Given a programming problem along with a code and a list of errors, your task is to:
1. Examine the code and all the errors in the list.
2. Reason which errors from the list are included in the code and return all that apply based on your reasoning.
3. Return an empty list if none of the errors are present in the code or the code is correct. \\ \\

The output MUST match this exact schema:\\
$\{$"errors": ["error 1", "error 2", ...]$\}$\\ \\

\textbf{User prompt:} \\
Problem: $\{$Problem Statement$\}$ \\ \\

Code: $\{$Code$\}$ \\ \\

Error list: $\{$Error list$\}$\\
\bottomrule
\end{tabular}
\caption{Prompt for LLM-as-a-Judge during the GRPO training loop and for human annotators during evaluation, ensuring consistent criteria across GRPO training and human assessment.}
\label{tab:human_eval}
\end{table*}

\section{Attribution}
Icons used in Figure~\ref{fig:model} were made by \href{https://www.flaticon.com/authors/uniconlabs}{uniconlabs}, \href{https://www.flaticon.com/authors/smashicons}{smashicons}, and \href{https://www.flaticon.com/authors/freepik}{freepik} from \url{www.flaticon.com}.

\end{document}